\documentclass[journal]{IEEEtran}
\usepackage[utf8]{inputenc}
\usepackage{cite}
\usepackage{hyperref}
\usepackage{multirow}

\usepackage{orcidlink}
\usepackage{subfigure}
\usepackage{amsmath}
\usepackage{amsfonts,amssymb}
\usepackage{colortbl}

\usepackage{makecell}
\usepackage{xcolor}
\usepackage{color}
\usepackage{threeparttable}
\usepackage{graphicx}
\usepackage{dblfloatfix}
\graphicspath{ {./figures/} }
\usepackage{tikz}
\usetikzlibrary{shapes.geometric}
\usepackage{microtype}

\hypersetup{
    colorlinks=true,
    linkcolor=cyan,
    filecolor=blue,
    urlcolor=black,
    citecolor=green,
}

\begin{document}

\title{A Novel Geographic Graph-Regulated Disentangling Mamba Model with Sparse Tokens for Enhanced Tree Species Classification from MODIS Time Series}

\author{Motasem Alkayid, Zhengsen Xu, Naser El-Sheimy,~\IEEEmembership{Member,~IEEE}, Lincoln Linlin Xu,~\IEEEmembership{Member,~IEEE} 
\thanks{This work was supported by the Natural Sciences and Engineering Research Council of Canada (NSERC) under Grant RGPIN-2019-06744.}
\thanks{Zhengsen Xu, Naser El-Sheimy, and Lincoln Linlin Xu are with the Department of Geomatics Engineering, University of Calgary, Canada (email: \{zhengsen.xu, elsheimy, lincoln.xu\}@ucalgary.ca) (Corresponding author: Lincoln Linlin Xu).}
\thanks{Motasem Alkayid is with the Department of Geomatics Engineering, University of Calgary, Canada, and also with the Department of Geography, Faculty of Arts, The University of Jordan, Amman, Jordan (email: motasem.alkayid@ucalgary.ca).}
}

\markboth{IEEE Geoscience and Remote Sensing Letters, March 2026}
{Alkayid \MakeLowercase{\textit{et al.}}: Geo-GDS-Mamba Tree Species Classification}
\maketitle

\begin{abstract}
Tree species classification from Moderate Resolution Imaging Spectroradiometer (MODIS) time series is critical for environmental monitoring, yet challenging due to subtle inter-species signatures, strong spatial-spectral-temporal coupling, and the difficulty of modeling large-scale geographical correlations. This letter presents a Graph-regulated Disentangled Sparse Mamba model (Geo-GDS-Mamba) with three contributions. First, a mini-batch graph-regulated stream models geographic correlations among samples via a Radial-Basis-Function kernel over their physical coordinates, independent of spectral content, and fuses this context through a graph convolutional network without full-batch memory costs. Second, a disentangling Mamba architecture separates the spatial, spectral, and temporal information coupled within each sample into three independent processing streams. Third, sparse token modules adaptively retain the top 30\% most informative tokens per branch via learned attention or similarity scores, mitigating Mamba's correlation-decay limitation while reducing computational cost. Experiments on annual MOD13Q1 data across two Canadian provinces show Geo-GDS-Mamba achieves 95.35\% OA (94.73\% Kappa) in Alberta, outperforming eleven state-of-the-art baselines, and generalizes without fine-tuning to unseen Saskatchewan data at 87.63\% OA, using only 1.74M parameters and 0.18 GFLOPs. Ablation studies confirm each component's contribution, led by a 4.15-point OA gain from the geographic graph stream alone.
\end{abstract}


\IEEEpeerreviewmaketitle
\vspace{-8pt}
\section{Introduction}

\begin{figure}[t]
    \centering
    \includegraphics[width=0.80\linewidth]{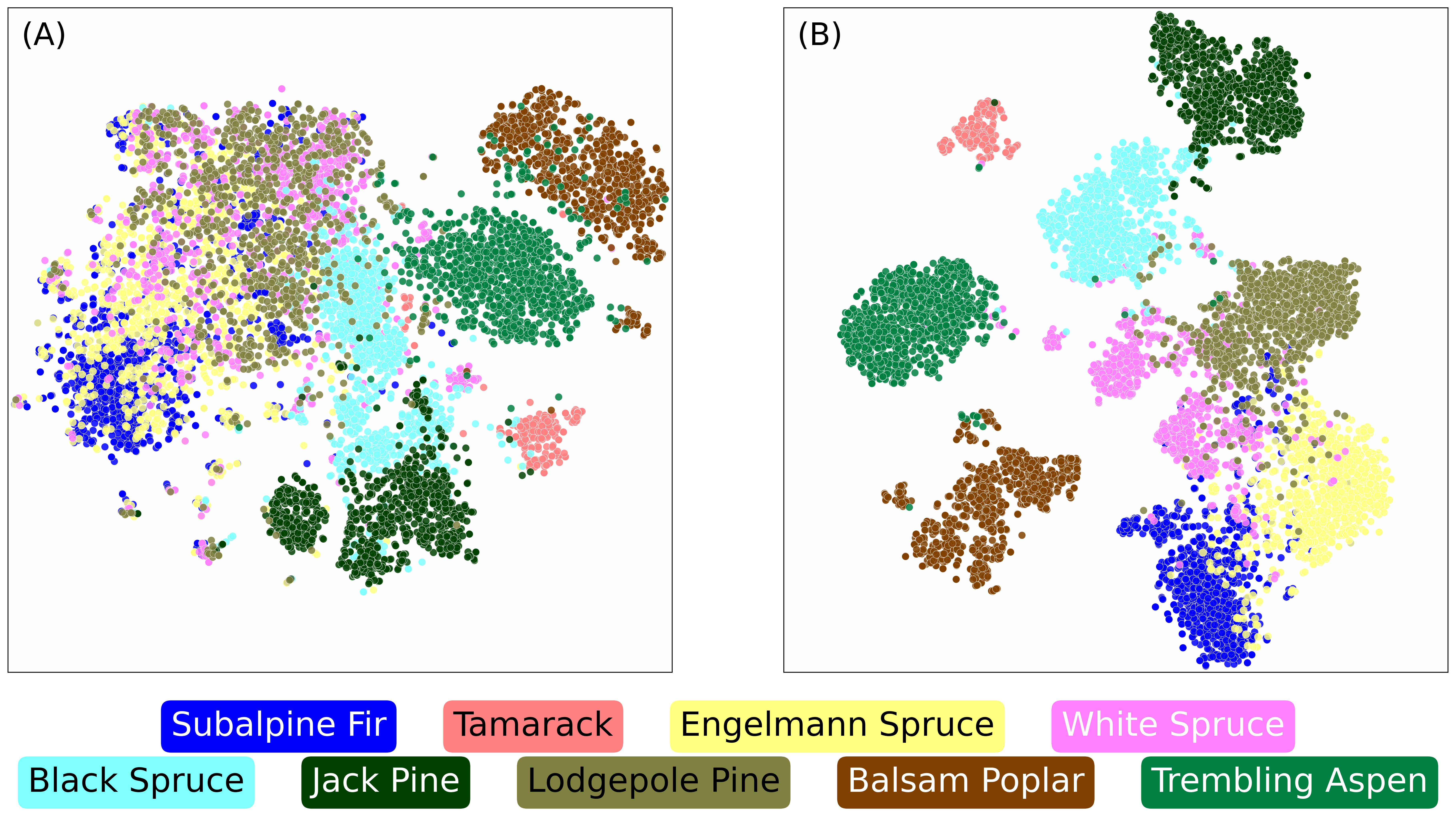}

    \vspace{-0.2cm} 

    \caption{t-SNE class separability of (A) raw data and (B) Geo-GDS-Mamba features for the Alberta 2010 dataset. The proposed Geo-GDS-Mamba model greatly improves MODIS tree species class separability and thereby the classification accuracy, because it systematically addresses three key issues that bottleneck standard Mamba models.}
    \label{fig:tnse_final}

\end{figure}

\begin{figure}[htbp]
    \vspace{-0.1cm} 
    \centering
    \includegraphics[width=0.40\textwidth]{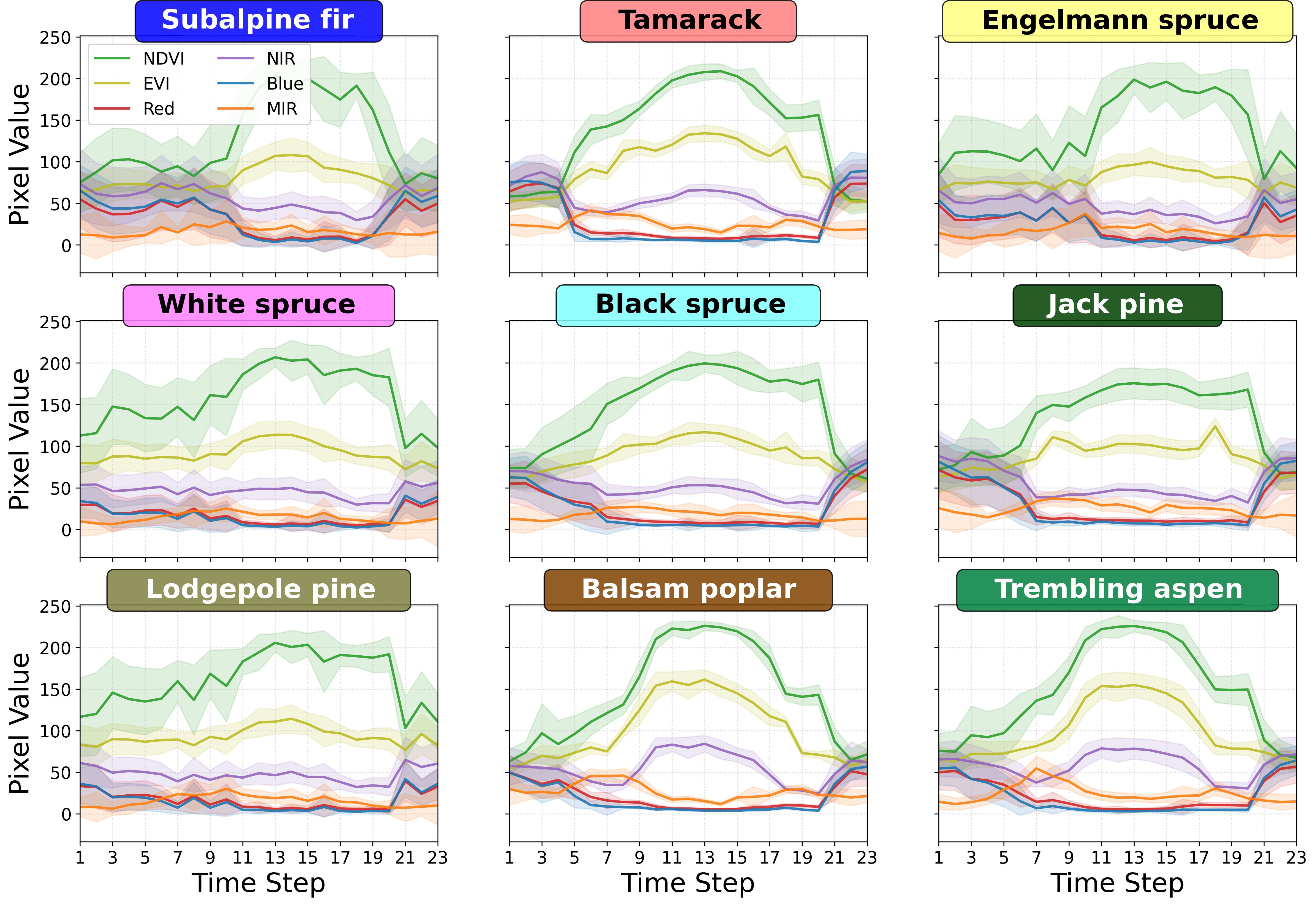}

    \vspace{-0.2cm} 

    \caption{The spectral-temporal curves of different classes in the Alberta dataset (23 time steps and 6 different spectral bands). The curves of most classes look similar and class signature differences are very subtle.}
    \label{fig:Spectral_Curve}
    \vspace{-0.1cm} 
\end{figure}

Accurate mapping of tree species over large, heterogeneous regions using MODIS time series is critical for environmental applications~\cite{hermosilla2024characterizing}. However, subtle species signatures embedded in high-dimensional spatial-spectral-temporal space (Fig.~\ref{fig:Spectral_Curve}) lead to challenging class inseparability (Fig.~\ref{fig:tnse_final}A), requiring models that disentangle this coupling and leverage large-scale geographical context.

Conventional deep learning architectures struggle to separate these patterns~\cite{shi2015convolutional}: CNNs are restricted by their locality inductive bias and fail to capture long-range dependencies; Transformers capture global dynamics, but their quadratic complexity deters application to MODIS's exceptionally long sequences~\cite{qin2025sitsmamba}; and GCNs~\cite{hong2020graph} model correlations between distant pixels, but full-batch processing of the entire graph introduces massive memory costs that limit scalability.

\begin{figure*}[tp]
    \centering
    \includegraphics[width=0.8\linewidth]{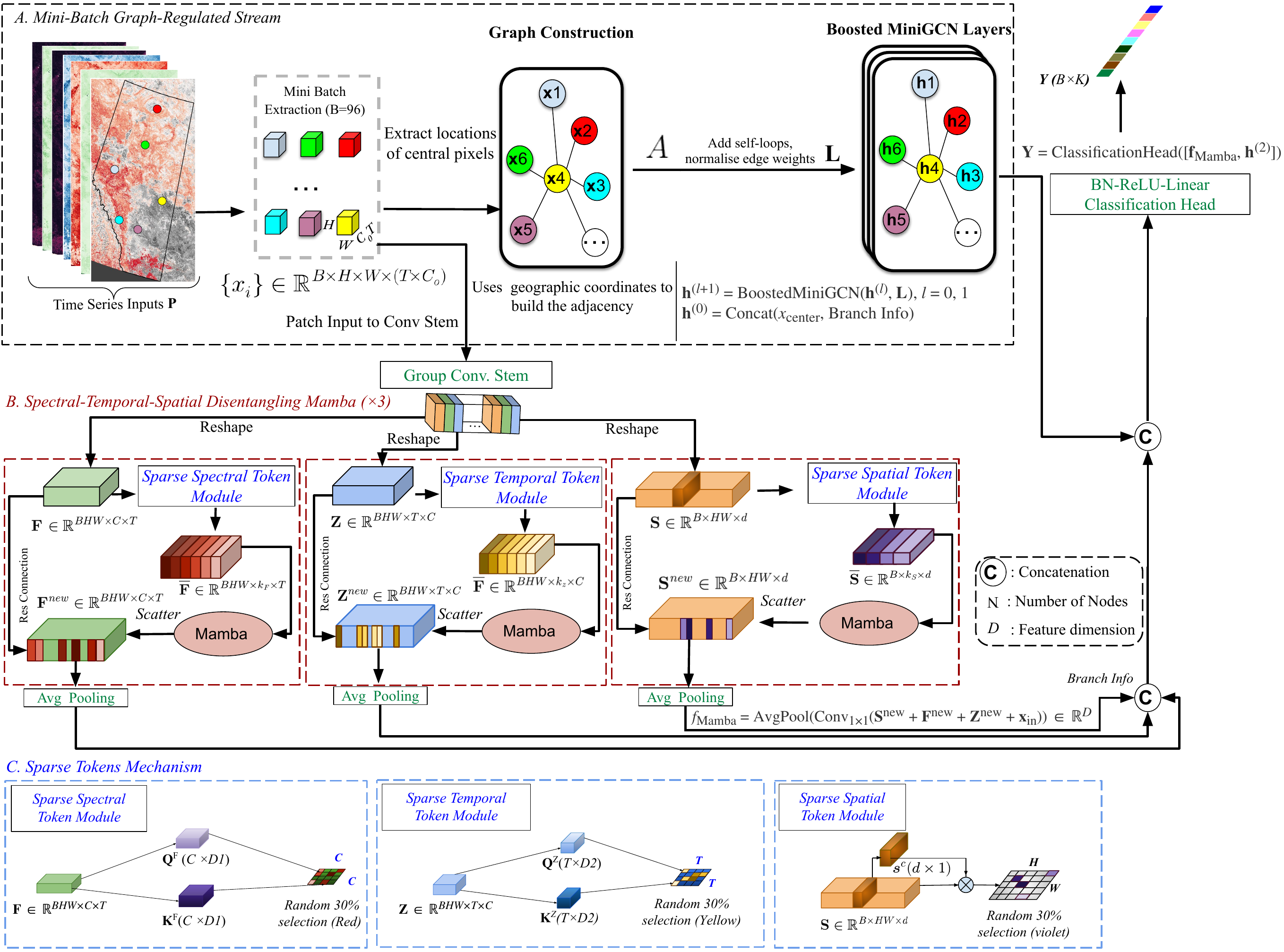}
    \caption{The architecture of the proposed model, with three key contributions. (1) First, the large-scale geographic spatial correlation within $\mathbf{P}$ is addressed using a mini-batch graph approach, where samples $\{\mathbf{x}_i \}$ are treated as nodes in a graph and modeled using the GCN method. (2) Second, the spatial-spectral-temporal coupling effect in MODIS time series data is addressed by the disentangling Mamba module, where spectral, temporal and spatial tokens, i.e., $\mathbf{F}$, $\mathbf{Z}$ and $\mathbf{S}$ respectively, are processed independently and fused with GCN features to generate labels $\mathbf{\textit{Y}}$. (3) Third, sparse token mechanism is used to address the key Mamba limitations, i.e., correlation decay and complexity with long token sequences.}

    \label{fig:model architecture}
\end{figure*}

More recently, structured state space sequence models (SSMs), i.e., Mamba, which model long-range correlations with reduced complexity, have been widely used in remote sensing~\cite{gu2023mamba,chen2024rsmamba}. However, standard Mamba cannot capture geographic correlation beyond individual input patches, does not separate the spatial, spectral, and temporal coupling intrinsic to MODIS time series, and its dense token sequencing suffers from the notorious correlation decay issue that bottlenecks performance~\cite{xu2025sparse}. Therefore, how to design geographically-guided disentangling Mamba models that simultaneously address these challenges is a critical research issue.

This letter therefore presents a novel Graph-regulated Disentangling Sparse Mamba (Geo-GDS-Mamba) framework for improved tree species classification from MODIS time series datasets, with the following contributions.

\begin{itemize}
\item First, to improve the modeling of large-scale geographic spatial correlation effect, we design a mini-batch graph regulated approach that explicitly explores geographic and physical spatial correlation among the input image patches.

\item Second, to disentangle the high-dimensional spatial-spectral-temporal information coupling effect in MODIS data for improved feature extraction, we propose a disentangling Mamba architecture tailored for capturing independent spatial pattern, spectral characteristics, and temporal phenology behaviors in MODIS time series.

\item Third, to improve efficiency and subtle feature learning, we design sparse token approaches that adaptively learn the informative subset of tokens to better address the correlation decay problem that bottlenecks standard Mamba models.
\end{itemize}

Fig.~\ref{fig:tnse_final} confirms this improved class separability.
\vspace{-8pt}

\section{Methodology}
\label{sec:methodology}
\vspace{-4pt}

\subsection{Mini-Batch Graph-Regulated Stream}
\label{sec:minigcn}
\vspace{-4pt}

According to Fig.~\ref{fig:model architecture}, the input large-scale MODIS time series image $\mathbf{P}$ is used to extract a mini-batch of 3D cubes, i.e.,  $\mathbf{M} =\{\boldsymbol{x}_1, \boldsymbol{x}_2, \dots, \boldsymbol{x}_B\}$, where $B$ is the mini-batch size and $\boldsymbol{x}_i \in \mathbb{R}^{H \times W \times TC_0}$, with $H$, $W$, $T$ and $C_0$ being respectively the height, width, number of time steps, and number of spectral features in the input 3D cube. Each sample $i$ also carries its physical spatial coordinate $\boldsymbol{g}_i \in \mathbb{R}^{2}$, the pixel row/column scaled by the 250\,m resolution.

To capture batch-level geographic correlations, we treat each sample in $\mathbf{M}$ as a graph node. The adjacency matrix $\mathbf{A} \in \mathbb{R}^{B \times B}$ is computed via a Radial Basis Function (RBF) kernel over geographic coordinates:
\begin{equation}
    A_{ij} = \exp\left(-\frac{\|\boldsymbol{g}_i - \boldsymbol{g}_j\|^2}{\sigma^2}\right)
    \label{eq:rbf_kernel}
\end{equation}
where $\sigma$ is a learnable bandwidth initialized at 10\,km, so that edges depend only on $\boldsymbol{g}_i$, while the node features enter as $\mathbf{H}^{(0)}$. We compute $\tilde{\mathbf{A}} = \mathbf{A} + \mathbf{I}$ with self-loops, and its diagonal degree matrix $\tilde{\mathbf{D}}$, where $\tilde{\mathbf{D}}_{i,i} = \sum_j \tilde{\mathbf{A}}_{i,j}$.

Based on $\tilde{\mathbf{A}}$ and $\tilde{\mathbf{D}}$, we compute the normalized adjacency matrix $\mathbf{L} = \tilde{\mathbf{D}}^{-1/2} \tilde{\mathbf{A}} \tilde{\mathbf{D}}^{-1/2}$. We apply a graph convolutional network (GCN) layer to learn new features $\mathbf{H}^{(l+1)}$ from input $\mathbf{H}^{(l)}$. The initial node features $\mathbf{H}^{(0)}$ are explicitly boosted by concatenating the raw spectral vectors with the intermediate spatial, spectral, and temporal Mamba branch features, i.e.,
\begin{equation}
    \mathbf{H}^{(l+1)} = \operatorname{BN}\left( \mathbf{L} \mathbf{H}^{(l)} \mathbf{W}^{(l)} + \boldsymbol{b}^{(l)} \right)
    \label{eq:gcn_propagation}
\end{equation}
where $\operatorname{BN}(\cdot)$ denotes batch normalization, with trainable parameters $\mathbf{W}^{(l)}$ and $\boldsymbol{b}^{(l)}$. The geographic proximity graph efficiently models inter-sample context across the mini-batch, while the parallel Mamba branches explicitly disentangle complex intra-sample dependencies.

\subsection{Spatial-Spectral-Temporal Disentangling Mamba}
\vspace{-4pt}

Fig.~\ref{fig:model architecture} indicates that a group convolution stem is used to generate the spectral tokens $\mathbf{F} \in \mathbb{R}^{BHW \times C \times T}$ with $C$ being the number of spectral tokens, i.e., $
    \mathbf{F} = \operatorname{groupConv}\left( \mathbf{M} \right)$,
where $\operatorname{groupConv}$ rearrange $\mathbf{M}$ into $T$ temporal groups and performs convolutions independently for each group to avoid mixing the temporal information with the spatial and spectral information.  Similarly, we obtain the temporal tokens $\mathbf{Z}_j \in \mathbb{R}^{BHW \times T \times C}$ with $T$ being the number of temporal tokens and the spatial tokens $\mathbf{S} \in \mathbb{R}^{B \times HW \times d}$ with $HW$ being the number of spatial tokens.

For spectral features $\mathbf{F} \in \mathbb{R}^{BHW \times C \times T}$, rather than using all $C$ tokens, we apply a binary selection mask $\mathbf{M}_F$ to extract a subset of $k_F$ top-scoring tokens ($k_F \ll C$), yielding $\bar{\mathbf{F}} \in \mathbb{R}^{BHW \times k_F \times T}$. Next, $\bar{\mathbf{F}}$ passes through the Mamba model to generate $k_F$ updated spectral tokens. These are scattered back to their original dimensions via the transposed mask $\mathbf{M}_F^{\top}$ to form $\mathbf{F}^{new}$:
\vspace{-1mm}
\begin{equation}
    \mathbf{F}^{new} = \mathbf{M}_F^{\top} \operatorname{Mamba}(\mathbf{M}_F \mathbf{F})
    \label{eq:spectral_scatter}
\end{equation}
\vspace{-5mm}

For temporal feature $\mathbf{Z}$ and spatial feature $\mathbf{S}$, we generate $\mathbf{Z}^{new}$, $\mathbf{S}^{new}$ using the Sparse Temporal Token Module and the Sparse Spatial Token Module respectively in a similar manner as $\mathbf{F}^{new}$.

The disentangling block is stacked three times, and the branch outputs of the last block are fused in two ways. First, they are summed with the block input $\mathbf{x}_{in}$, passed through a $1{\times}1$ convolution, and globally average-pooled, i.e., $\mathbf{f}_{\mathrm{Mamba}} = \operatorname{AvgPool}\big(\operatorname{Conv}_{1\times1}(\mathbf{S}^{new} + \mathbf{F}^{new} + \mathbf{Z}^{new} + \mathbf{x}_{in})\big) \in \mathbb{R}^{D}$. Second, each branch is average-pooled separately and concatenated into the branch information $\mathbf{b} = [\operatorname{AvgPool}(\mathbf{S}^{new}) \,\|\, \operatorname{AvgPool}(\mathbf{F}^{new}) \,\|\, \operatorname{AvgPool}(\mathbf{Z}^{new})] \in \mathbb{R}^{3D}$, which forms the GCN input $\mathbf{H}^{(0)} = [\mathbf{x}_{center} \,\|\, \mathbf{b}]$ (Section~\ref{sec:minigcn}). Finally, $\mathbf{f}_{\mathrm{Mamba}}$ is concatenated with the output of the last GCN layer, $\mathbf{H}^{(2)}$, to generate the output labels $\mathbf{Y} (B \times K) = \operatorname{ClassificationHead}([\mathbf{f}_{\mathrm{Mamba}} \,\|\, \mathbf{H}^{(2)}])$, where $K$ is the number of classes and the classification head consists of batch normalization, ReLU, dropout, and a linear layer.

\subsection{Sparse Tokens Mechanism}
\label{sec:sparse}
\vspace{-4pt}

The Sparse Spectral Token Module and Sparse Temporal Token Module in Fig.~\ref{fig:model architecture} score each token with learnable query--key projections. For a normalized token sequence $\hat{\mathbf{X}} = [\hat{\mathbf{x}}_1, \dots, \hat{\mathbf{x}}_L]$, with $L = C$ for $\mathbf{F}$ and $L = T$ for $\mathbf{Z}$, the importance score of the $i$th token is its scaled query--key affinity averaged over all tokens and heads:
\begin{equation}
    \gamma_i = \frac{1}{H_{\text{heads}}\, L} \sum_{h=1}^{H_{\text{heads}}} \sum_{j=1}^{L} \frac{\mathbf{q}_{h,i}^{\top}\, \mathbf{k}_{h,j}}{\sqrt{d_k}}
    \label{eq:spectral_temporal_attn}
\end{equation}
\noindent where $\mathbf{q}_{h,i} = \mathbf{W}_Q^{h} \hat{\mathbf{x}}_i$ and $\mathbf{k}_{h,j} = \mathbf{W}_K^{h} \hat{\mathbf{x}}_j$ are the query of token $i$ and the key of token $j$ in head $h$, $\mathbf{W}_Q^{h}$ and $\mathbf{W}_K^{h}$ are learnable projections, $H_{\text{heads}}$ is the number of heads, and $d_k$ is the key dimension, used to scale the logits and stabilize the gradients during training. Thus, $\gamma_i$ measures how strongly the query of token $i$ aligns with the keys of the whole sequence, i.e., its relational centrality rather than its activation magnitude.

The Sparse Spatial Token Module scores each spatial token $\boldsymbol{s}_i$ by its cosine similarity to the central token $\boldsymbol{s}_c$ of the patch:
\begin{equation}
    \gamma_i = \frac{\boldsymbol{s}_i^{\top}\, \boldsymbol{s}_c}{\|\boldsymbol{s}_i\|\;\|\boldsymbol{s}_c\|}
    \label{eq:spatial_attn}
\end{equation}
\noindent where $\boldsymbol{s}_i$ are the learnably projected spatial tokens. The central token is used as the reference because it is the prediction target, and the L2-normalised score is magnitude-free.

In each branch, the $k = \lfloor 0.3L \rfloor$ highest-scoring tokens form the selected set $\mathcal{I}$, i.e., $k_F = \lfloor 0.3C \rfloor$, $k_Z = \lfloor 0.3T \rfloor$, and $k_s = \lfloor 0.3HW \rfloor$. As top-$k$ selection is not differentiable, each selected token is scaled by a sigmoid gate of its score, $\bar{\mathbf{x}}_i = \sigma(\gamma_i)\,\hat{\mathbf{x}}_i$ for $i \in \mathcal{I}$, i.e., $\mathbf{G} = \operatorname{diag}(\sigma(\gamma_i))$ in Eq.~\eqref{eq:spectral_scatter}, which passes the loss gradient to the scoring projections, so the selection is learned end to end.

Mamba reads tokens in order and carries a hidden state $\boldsymbol{h}_t$ forward. At every step, this state is multiplied by $\mathbf{A}$, which shrinks it, so the effect of an input $\boldsymbol{x}_{t-k}$ from $k$ steps earlier fades as $\mathbf{A}^{k}$~\cite{gu2022s4}. This fading, called correlation decay, motivated the selective design of Mamba~\cite{gu2023mamba}. Keeping only the top-$k$ tokens makes the sequence shorter, so useful tokens stay closer and fade less; the skipped tokens are not lost, as they go around the Mamba through the residual connection.

\vspace{-12pt}

\subsection{Experimental Settings}
\vspace{-4pt}
We used MODIS (MOD13Q1) 2010 time-series data with 23 temporal steps and six variables (B01, B02, B03, B07, NDVI, EVI) at 250m resolution, extracting $13{\times}13$ patches. Ground truth labels were derived from 30m Canadian Tree Species maps~\cite{hermosilla2024characterizing} (nine classes in Alberta, four in Saskatchewan), aggregated to 250m and cleaned with a $7{\times}7$ majority filter retaining only pixels where at least 80\% of the surrounding window shares the same class. To prevent spatial data leakage, we used a spatially disjoint $15{\times}15$ block holdout with per-class caps of 100/100/1,000 for train/val/test. The network used Adam (lr=$1{\times}10^{-3}$, weight decay=$1{\times}10^{-4}$) with batch size 96 for 100 epochs with early stopping. The Mamba hidden dimension was 276 and GCN output dimension 276.

The Saskatchewan dataset evaluated cross-provincial generalization without fine-tuning or re-labelling. While Alberta has 9 classes, Saskatchewan contains only 4 (Tamarack, Black spruce, Jack pine, Trembling aspen). Species discrimination leverages phenological timing across the 23 temporal steps: deciduous species exhibit distinct seasonal NDVI/EVI trajectories absent in evergreen conifers, enabling the Temporal branch to separate spectrally overlapping classes (Fig.~\ref{fig:Spectral_Curve}). Among conifers, canopy-structure differences produce distinct NIR and MIR responses exploited by the Spectral branch, while the Spatial branch resolves mixed-stand ambiguity at stand boundaries. We compare against eleven state-of-the-art baselines: ConvLSTM~\cite{shi2015convolutional}, SiT~\cite{chen2025sit}, ResNet-101~\cite{7780459}, ConvNeXt~\cite{liu2022convnet}, SSEFN~\cite{liu2025ssefn}, SwinT~\cite{liu2021swin}, Mamba-HSI~\cite{zhao2024mambahsi}, RS-Mamba~\cite{chen2024rsmamba}, GraphMamba~\cite{yang2024graphmamba}, UTAE~\cite{garnot2021panoptic}, and SITSMamba~\cite{qin2025sitsmamba}. All models used center-pixel targets with weighted cross-entropy loss, and results are averaged across three random seeds.
\vspace{-10pt}

\subsection{Comparison Results}
\begin{figure*}[h]
    \centering
    \includegraphics[width=\textwidth]{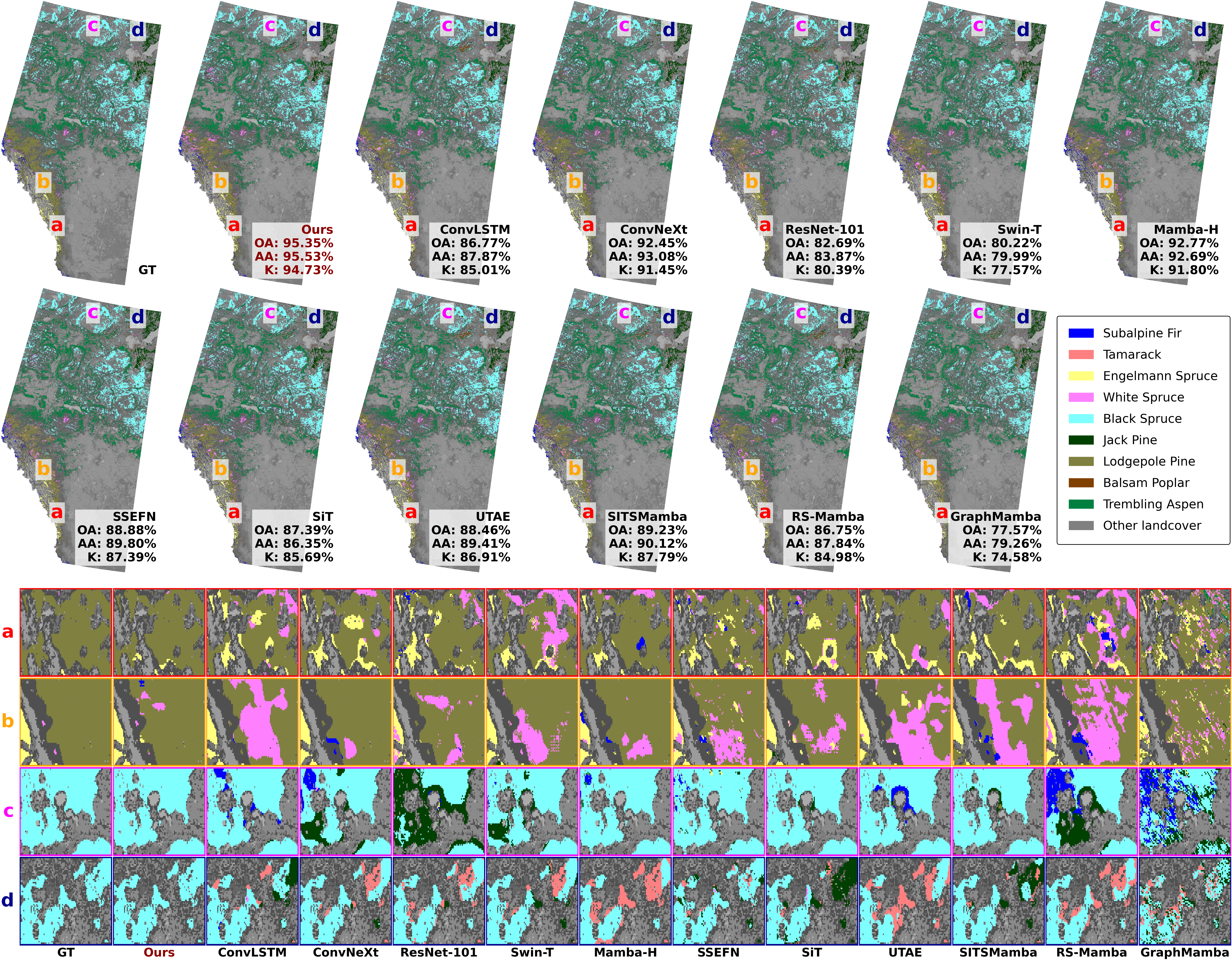}
    \caption{Classification maps achieved by different methods on the full filtered Alberta map. The zoom in views of four regions (i.e., a, b, c, and d) indicate that our approach achieves fine-grained details that are the most consistent with the ground-truth (GT). OA/AA/K values are the test-set results from Table~\ref{Alberta2010}.}
    \label{fig:alberta_map}
\end{figure*}

\vspace{-4pt}
Table~\ref{Alberta2010} shows that our model achieves $95.35\%\pm0.03\%$ OA across three random seeds on the disjoint spatial block holdout, outperforming all eleven baselines, the strongest of which are Mamba-HSI (92.77\%) and ConvNeXt (92.45\%). Our model is slightly lower than Mamba-HSI on White-spruce (87.70\% vs.\ 91.60\%), as its single-stream spectral processing favours classes with highly distinctive spectral profiles. However, the disentangling structure benefits the majority of species that share overlapping spectral signatures but differ in phenological timing, yielding a superior average accuracy (95.53\% vs.\ 92.69\%) across all nine classes. Fig.~\ref{fig:alberta_map} shows that, on the full filtered Alberta map, our model yields the map most consistent with the ground truth, giving sharper stand boundaries and the least overestimation among all baselines.

In cross-provincial transfer to Saskatchewan (Table~\ref{SaskatchewanTestPerformance}), our model attains the highest OA (87.63\%), AA (85.99\%), and Kappa (82.26\%) without any fine-tuning. The lower OA relative to Alberta reflects the reduced species set (4 of 9 classes) and cross-provincial domain shift; per class, our model leads on Trembling-aspen (99.90\%) while remaining competitive on Jack-pine (96.50\% vs.\ 96.60\% for SiT).

\begin{table}[b]
\centering
\small 
\setlength{\tabcolsep}{6pt} 
\renewcommand{\arraystretch}{1.2} 
\caption{Ablation Study: Feature Branch and Graph Construction.}
\vspace{-2mm}
\resizebox{\linewidth}{!}{%
\begin{tabular}{l c c c c}
\hline
\multirow{2}{*}{\textbf{Configuration / Variant}} & \multicolumn{2}{c}{\textbf{Alberta}} & \multicolumn{2}{c}{\textbf{Saskatchewan}} \\
 & OA (\%) & Kappa & OA (\%) & Kappa \\ \hline
\rowcolor[RGB]{251,228,213} \textbf{Proposed: Geo-GDS-Mamba} & \textbf{95.35 $\pm$ 0.03} & \textbf{0.9473 $\pm$ 0.0003} & \textbf{87.63 $\pm$ 0.64} & \textbf{0.8226 $\pm$ 0.0094} \\ \hline
\multicolumn{5}{l}{\textit{\textbf{1. Feature Branch Ablations}}} \\
w/o Spatial Branch & 93.45 $\pm$ 0.55 & 0.9257 $\pm$ 0.0063 & 86.15 $\pm$ 0.34 & 0.8019 $\pm$ 0.0053 \\
w/o Spectral Branch & 93.81 $\pm$ 0.67 & 0.9298 $\pm$ 0.0076 & 86.13 $\pm$ 0.61 & 0.8016 $\pm$ 0.0089 \\
w/o Temporal Branch & 93.64 $\pm$ 0.46 & 0.9278 $\pm$ 0.0053 & 86.84 $\pm$ 0.40 & 0.8113 $\pm$ 0.0058 \\ \hline
\multicolumn{5}{l}{\textit{\textbf{2. Graph Construction Ablations}}} \\
Hybrid Geo-Spectral Graph & 93.93 $\pm$ 0.45 & 0.9324 $\pm$ 0.0044 & 86.91 $\pm$ 0.19 & 0.8129 $\pm$ 0.0018 \\
w/o Graph & 91.20 $\pm$ 0.21 & 0.9002 $\pm$ 0.0024 & 85.87 $\pm$ 0.24 & 0.7981 $\pm$ 0.0036 \\
w/o Mamba-GCN Residual & 91.56 $\pm$ 0.25 & 0.9043 $\pm$ 0.0029 & 86.71 $\pm$ 0.04 & 0.8100 $\pm$ 0.0007 \\ \hline
\end{tabular}%
}
\label{tab:ablation1}
\end{table}

\begin{table}[b]
\centering
\small 
\setlength{\tabcolsep}{6pt} 
\renewcommand{\arraystretch}{1.2} 
\caption{Ablation Study: Fusion Strategy and Token Sparsity.}
\vspace{-2mm}
\resizebox{\linewidth}{!}{%
\begin{tabular}{l c c c c}
\hline
\multirow{2}{*}{\textbf{Configuration / Variant}} & \multicolumn{2}{c}{\textbf{Alberta}} & \multicolumn{2}{c}{\textbf{Saskatchewan}} \\
 & OA (\%) & Kappa & OA (\%) & Kappa \\ \hline
\rowcolor[RGB]{251,228,213} \textbf{Proposed: Geo-GDS-Mamba} & \textbf{95.35 $\pm$ 0.03} & \textbf{0.9473 $\pm$ 0.0003} & \textbf{87.63 $\pm$ 0.64} & \textbf{0.8226 $\pm$ 0.0094} \\ \hline
\multicolumn{5}{l}{\textit{\textbf{3. Fusion Strategy Ablations (vs.\ concatenation)}}} \\
Fusion: Additive & 93.15 $\pm$ 0.30 & 0.9223 $\pm$ 0.0035 & 86.32 $\pm$ 0.60 & 0.8046 $\pm$ 0.0089 \\
Fusion: GLU & 93.33 $\pm$ 0.44 & 0.9244 $\pm$ 0.0050 & 86.80 $\pm$ 1.38 & 0.8114 $\pm$ 0.0194 \\
Fusion: Cross-Attention & 90.80 $\pm$ 0.20 & 0.8957 $\pm$ 0.0023 & 85.39 $\pm$ 0.84 & 0.7928 $\pm$ 0.0108 \\
Fusion: SE & 92.60 $\pm$ 1.02 & 0.9161 $\pm$ 0.0115 & 86.05 $\pm$ 1.44 & 0.8011 $\pm$ 0.0201 \\
Fusion: FiLM & 93.00 $\pm$ 0.62 & 0.9206 $\pm$ 0.0071 & 86.25 $\pm$ 0.11 & 0.8033 $\pm$ 0.0013 \\ \hline
\multicolumn{5}{l}{\textit{\textbf{4. Token Threshold \& Sparsity Ablations}}} \\
w/o Token Sparsity & 93.70 $\pm$ 0.66 & 0.9285 $\pm$ 0.0075 & 85.61 $\pm$ 0.80 & 0.7942 $\pm$ 0.0112 \\
Deterministic (non-learned) Top-30\% & 93.21 $\pm$ 0.15 & 0.9229 $\pm$ 0.0017 & 84.49 $\pm$ 2.60 & 0.7800 $\pm$ 0.0354 \\ \hline
\end{tabular}%
}
\label{tab:ablation2}
\end{table}

\begin{table*}[!h]
\centering
\footnotesize 
\renewcommand{\arraystretch}{0.7} 
\setlength{\tabcolsep}{3pt}
\caption{\scriptsize Classification results on the Alberta 2010 test dataset. The best results are in bold with color shadow.}
\vspace{-1mm}
\resizebox{\textwidth}{!}{%
\begin{tabular}{cc|ccccccc|cccc|c}
\hline
\multicolumn{1}{c|}{Colour} & \multicolumn{1}{c|}{Class Name} & ConvLSTM & SIT & ResNet-101 & ConvNeXt & SSEFN & SwinT & Mamba-HSI & RS-Mamba & GraphMamba & UTAE & SITSMamba & Ours (Geo-GDS-Mamba) \\ \hline
\cellcolor[RGB]{0, 0, 255} & \multicolumn{1}{c|}{Subalpine-fir} & 76.70 & 85.00 & 74.60 & 90.80 & 84.80 & 77.00 & \underline{92.20} & 78.20 & 68.20 & 81.40 & 83.60 & \cellcolor[RGB]{251, 228, 213}\textbf{95.00} \\
\cellcolor[RGB]{255, 128, 128} & \multicolumn{1}{c|}{Tamarack} & \cellcolor[RGB]{251, 228, 213}\textbf{100.00} & 74.71 & 96.89 & \cellcolor[RGB]{251, 228, 213}\textbf{100.00} & \cellcolor[RGB]{251, 228, 213}\textbf{100.00} & 77.43 & 91.83 & \cellcolor[RGB]{251, 228, 213}\textbf{100.00} & \underline{98.05} & \cellcolor[RGB]{251, 228, 213}\textbf{100.00} & \cellcolor[RGB]{251, 228, 213}\textbf{100.00} & \cellcolor[RGB]{251, 228, 213}\textbf{100.00} \\
\cellcolor[RGB]{255, 255, 128} & \multicolumn{1}{c|}{Engelmann-spruce} & \underline{89.50} & 84.30 & 69.90 & 86.20 & 82.40 & 81.60 & 85.00 & 82.50 & 63.30 & \underline{89.50} & 85.60 & \cellcolor[RGB]{251, 228, 213}\textbf{91.50} \\
\cellcolor[RGB]{255, 128, 255} & \multicolumn{1}{c|}{White-spruce} & 71.40 & \underline{90.40} & 64.10 & 89.30 & 73.50 & 62.30 & \cellcolor[RGB]{251, 228, 213}\textbf{91.60} & 73.50 & 41.90 & 66.50 & 78.70 & 87.70 \\
\cellcolor[RGB]{128, 255, 255} & \multicolumn{1}{c|}{Black-spruce} & 85.80 & 79.60 & 87.50 & 90.80 & \underline{94.20} & 54.60 & 92.30 & 91.90 & 83.10 & 90.10 & 90.20 & \cellcolor[RGB]{251, 228, 213}\textbf{95.80} \\
\cellcolor[RGB]{0, 64, 0} & \multicolumn{1}{c|}{Jack-pine} & 94.90 & 97.40 & 94.00 & \underline{98.40} & 96.80 & 97.30 & 98.30 & 94.70 & 93.30 & 98.00 & 96.70 & \cellcolor[RGB]{251, 228, 213}\textbf{98.70} \\
\cellcolor[RGB]{128, 128, 64} & \multicolumn{1}{c|}{Lodgepole-pine} & 76.40 & 77.20 & 75.60 & \underline{92.20} & 80.80 & 80.60 & 89.50 & 77.30 & 73.90 & 82.40 & 80.80 & \cellcolor[RGB]{251, 228, 213}\textbf{93.40} \\
\cellcolor[RGB]{128, 64, 0} & \multicolumn{1}{c|}{Balsam-poplar} & 99.80 & 99.80 & \underline{99.90} & \underline{99.90} & \cellcolor[RGB]{251, 228, 213}\textbf{100.00} & \cellcolor[RGB]{251, 228, 213}\textbf{100.00} & 99.50 & \cellcolor[RGB]{251, 228, 213}\textbf{100.00} & 99.20 & \underline{99.90} & \underline{99.90} & \cellcolor[RGB]{251, 228, 213}\textbf{100.00} \\
\cellcolor[RGB]{0, 128, 64} & \multicolumn{1}{c|}{Trembling-aspen} & 96.30 & 88.70 & 92.30 & 90.10 & 95.70 & 89.10 & 94.00 & 92.50 & 92.40 & \underline{96.90} & 95.60 & \cellcolor[RGB]{251, 228, 213}\textbf{97.70} \\ \hline
\multicolumn{2}{c|}{OA(\%)} & 86.77 & 87.39 & 82.69 & 92.45 & 88.88 & 80.22 & \underline{92.77} & 86.75 & 77.57 & 88.46 & 89.23 & \cellcolor[RGB]{251, 228, 213}\textbf{95.35} \\
\multicolumn{2}{c|}{AA(\%)} & 87.87 & 86.35 & 83.87 & \underline{93.08} & 89.80 & 79.99 & 92.69 & 87.84 & 79.26 & 89.41 & 90.12 & \cellcolor[RGB]{251, 228, 213}\textbf{95.53} \\
\multicolumn{2}{c|}{Kappa(\%)} & 85.01 & 85.69 & 80.39 & 91.45 & 87.39 & 77.57 & \underline{91.80} & 84.98 & 74.58 & 86.91 & 87.79 & \cellcolor[RGB]{251, 228, 213}\textbf{94.73} \\ \hline
\end{tabular}%
}
\label{Alberta2010}
\vspace{-4pt}
\end{table*}
\vspace{-6pt}

\begin{table*}[tp]
\centering
\footnotesize 
\renewcommand{\arraystretch}{0.7} 
\setlength{\tabcolsep}{3pt}
\caption{\scriptsize Cross provincial generalization on the Saskatchewan test set. The best results are in bold with color shadow.}
\vspace{-1mm}
\resizebox{\textwidth}{!}{%
\begin{tabular}{cc|ccccccc|cccc|c}
\hline
\multicolumn{1}{c|}{Colour} & \multicolumn{1}{c|}{Class Name} & ConvLSTM & SIT & ResNet-101 & ConvNeXt & SSEFN & SwinT & Mamba-HSI & RS-Mamba & GraphMamba & UTAE & SITSMamba & Ours (Geo-GDS-Mamba) \\ \hline
\cellcolor[RGB]{255, 128, 128} & \multicolumn{1}{c|}{Tamarack} & \cellcolor[RGB]{251, 228, 213}\textbf{78.53} & 71.78 & 75.46 & 71.78 & \underline{76.07} & \cellcolor[RGB]{251, 228, 213}\textbf{78.53} & 71.78 & 74.85 & 72.39 & 71.78 & 71.78 & 71.78 \\
\cellcolor[RGB]{128, 255, 255} & \multicolumn{1}{c|}{Black-spruce} & 66.90 & 54.20 & 52.60 & 67.20 & \cellcolor[RGB]{251, 228, 213}\textbf{76.20} & 25.40 & 64.30 & 64.30 & 60.40 & 50.10 & 61.70 & \underline{75.78} \\
\cellcolor[RGB]{0, 64, 0} & \multicolumn{1}{c|}{Jack-pine} & 87.20 & \cellcolor[RGB]{251, 228, 213}\textbf{96.60} & 89.60 & 90.30 & 85.10 & 93.70 & 90.80 & 80.20 & 83.20 & 95.40 & 92.40 & \underline{96.50} \\
\cellcolor[RGB]{64, 128, 0} & \multicolumn{1}{c|}{Trembling-aspen} & 98.10 & 97.30 & 94.40 & 95.80 & 98.30 & 94.80 & 97.20 & 98.20 & 98.10 & \underline{99.70} & 98.90 & \cellcolor[RGB]{251, 228, 213}\textbf{99.90} \\
\hline
\multicolumn{2}{c|}{OA(\%)} & 83.78 & 82.14 & 78.69 & 83.78 & \underline{85.99} & 71.67 & 83.47 & 80.59 & 80.15 & 81.22 & 83.69 & \cellcolor[RGB]{251, 228, 213}\textbf{87.63} \\
\multicolumn{2}{c|}{AA(\%)} & 82.68 & 79.97 & 78.02 & 81.27 & \underline{83.92} & 73.11 & 81.02 & 79.39 & 78.52 & 79.24 & 81.19 & \cellcolor[RGB]{251, 228, 213}\textbf{85.99} \\
\multicolumn{2}{c|}{Kappa(\%)} & 77.02 & 74.56 & 70.04 & 76.89 & \underline{79.96} & 61.45 & 76.74 & 72.45 & 72.09 & 73.33 & 76.76 & \cellcolor[RGB]{251, 228, 213}\textbf{82.26} \\
\hline
\end{tabular}%
}
\label{SaskatchewanTestPerformance}
\vspace{-4pt}
\end{table*}
\vspace{-6pt}

\begin{table}[htbp]
\centering
\caption{Model Complexity Comparison for Lightweight Architectures.}
\vspace{-5pt}
\scriptsize
\resizebox{0.48\textwidth}{!}{
\begin{tabular}{lccc}
\hline
\textbf{Model} & \textbf{Parameters (M)} & \textbf{FLOPs (G)} & \textbf{Alberta OA (\%)} \\ \hline
ConvNeXt & 0.71 & 0.24 & 92.45 \\
UTAE & 1.02 & 2.37 & 88.46 \\
SITS-Mamba & 0.54 & 1.54 & 89.23 \\
Mamba-HSI & 1.83 & 0.24 & 92.77 \\
\rowcolor[RGB]{251, 228, 213} \textbf{Ours (Geo-GDS-Mamba)} & \textbf{1.74} & \textbf{0.18} & \textbf{95.35} \\\hline
\end{tabular}
}
\label{tab:flops_params}
\end{table}

\subsection{Computational Efficiency}
\vspace{-4pt}
Table~\ref{tab:flops_params} compares model complexity. With only 1.74M parameters and 0.18 GFLOPs, our model achieves the highest accuracy while being ${\sim}10\times$ faster than SITS-Mamba (1.54 GFLOPs) and ${\sim}600\times$ faster than ConvLSTM (107 GFLOPs). Our pure geographic-distance graph bypasses the dense matrix multiplications that bottleneck typical GCNs, making it well-suited for province-wide deployment.

\subsection{Ablation Studies}
\vspace{-4pt}
Table~\ref{tab:ablation1} shows that removing Spatial or Temporal branches causes the largest drops (1.90\% and 1.71\% OA on Alberta), confirming their critical role. The Graph module contributes a 4.15\% OA improvement. Table~\ref{tab:ablation2} shows that, for sparsity, retaining 30\% of tokens yields optimal performance; disabling sparsity or using non-learned top-$k$ selection both degrade accuracy due to Mamba's correlation decay. The concatenation fusion used in the proposed model outperforms all tested alternatives (additive, GLU, cross-attention, SE, and FiLM). To directly verify the disentanglement rather than assert it, a Pearson correlation probing analysis on the block-holdout test set (3 seeds) shows that, after removing the shared residual skip connection, the inter-branch feature correlations are all $|r|\leq0.02$, confirming that the three branches learn mutually uncorrelated representations.
\section{Conclusion} \label{conclusion}
This letter presented Geo-GDS-Mamba for tree species classification from MODIS time series, contributing: (1) a mini-batch geographic graph for large-scale spatial correlations; (2) a disentangling Mamba for independent spatial-spectral-temporal feature extraction; and (3) sparse token selection addressing Mamba's correlation decay. Experiments on Alberta and Saskatchewan demonstrated state-of-the-art accuracy with strong cross-regional generalization.

\vspace{-8pt}
\bibliographystyle{IEEEtran}
\bibliography{references}

\end{document}